\documentclass{article}
\usepackage[utf8]{inputenc}

\usepackage[normalem]{ulem}

\usepackage{nicematrix}

\usepackage{tikz}
\usetikzlibrary{positioning}

\usepackage[a4paper,top=2cm,bottom=2cm,left=2.5cm,right=2.5cm,marginparwidth=2cm]{geometry}

\usepackage{hyperref}
\hypersetup{
    colorlinks=true,
    linkcolor=blue,
    filecolor=magenta,      
    urlcolor=cyan,
    pdftitle={Overleaf Example},
    pdfpagemode=FullScreen,
    }

\usepackage{algpseudocode}

\usepackage{hyperref}
\usepackage{array}
\usepackage{mdframed}
\usepackage{mathtools}

\usepackage{verbatim}

\usepackage{tikz}
\usetikzlibrary{positioning}
\usetikzlibrary{arrows}
\usetikzlibrary{shapes.multipart}

\usepackage{booktabs}

\usepackage{amsthm}
\usepackage{amssymb}
\usepackage{amsmath, ulem}
\usepackage{blkarray}
\usepackage{graphicx}
\usepackage{multicol}

\usepackage{xcolor}

\usepackage{caption}
\usepackage{subcaption}

\newcommand{\stkout}[1]{\ifmmode\text{\sout{\ensuremath{#1}}}\else\sout{#1}\fi}

\usepackage{multirow}
\usepackage{soul}

\usepackage{textcomp}

\usepackage{accents}

\usepackage{algorithm}

\DeclareMathOperator*{\argmin}{argmin}

\usepackage{csquotes}

\newcommand{\suchthat}{\;\ifnum\currentgrouptype=16 \middle\fi|\;}

\newcommand\R{\mathbb{R}}

\newcommand\calG{\mathcal G}
\newcommand\calH{\mathcal H}

\DeclareMathAlphabet{\mathpzc}{OT1}{pzc}{m}{it}

\newcounter{casenum}

\definecolor{mygreen}{rgb}{0.0,0.7,0.0}
\definecolor{mybrown}{rgb}{0.5,0.5,0.0}

\title{
{\huge \textbf{Countering adversarial evasion\\ in regression analysis}}}
\author{\\[1ex]
David Benfield, Phan Tu Vuong, and Alain Zemkoho\\[1.5ex]
School of Mathematical Sciences\\ University of Southampton\\  SO17 1BJ Southampton, United Kingdom \\
\href{db3g17@soton.ac.uk}{db3g17@soton.ac.uk}, \href{t.v.phan@soton.ac.uk}{t.v.phan@soton.ac.uk}, \href{a.b.zemkoho@soton.ac.uk}{a.b.zemkoho@soton.ac.uk}\\[1.5ex]}
\date{}

\begin{document}


\maketitle


\begin{abstract}
Adversarial machine learning challenges the assumption that the underlying distribution remains consistent throughout the training and implementation of a prediction model. In particular, adversarial evasion considers scenarios where adversaries adapt their data to influence particular outcomes from established prediction models; such scenarios arise in applications such as spam email filtering, malware detection and fake-image generation, where security methods must be actively updated to keep up with the ever-improving generation of malicious data. Game theoretic models have been shown to be effective at modelling these scenarios and hence training resilient predictors against such adversaries. In this paper, we introduce a pessimistic bilevel optimsiation model, based on Stackelberg leader-follower games, to counter adversarial evasion in regression analysis. Unlike in some existing literature, we do not make assumptions (such as lower-level convexity or uniqueness of optimal solutions) on the adversary's optimal strategy to avoid hindering their capacity and hence capture their antagonistic nature, leading to more resilient prediction models. Further to this, through the introduction of lower-level constraints, we measure and restrict the adversary's movement, which, unlike its unrestricted counterpart in the current literature, prevents drastic transformations that do not accurately represent reality.

\end{abstract}




\section{Introduction}
\label{sctn:Introduction}

Adversarial machine learning considers the exploitable vulnerabilities of machine learning models and the strategies needed to counter or mitigate such threats \cite{adv_book_2023}. By considering these vulnerabilities during the development stage of our machine learning models, we can work to build resilient methods \cite{adv_deep_learning_survey, GameTheorySurvey} such as protection from credit card fraud \cite{INFORMS_credit_card_fraud_2} or finding the optimal placement of air defence systems \cite{INFORMS_defense}. In particular, we consider the model's sensitivity to changes in the distribution of the data. The way the adversary influences the distribution can fall under numerous categories, see \cite{taxonomy} for a helpful taxonomy that categorises these attacks. We focus on the specific case of evasion attacks, which consider the scenarios where adversaries attempt to modify their data to influence particular outcomes from prediction models. Such attacks might occur in security scenarios such as malware detection \cite{Malware_Detection} and network intrusion traffic \cite{Network_Intrusion}. In a similar vein, and more recently, vulnerabilities in deep neural networks (DNN) are being discovered, particularly in the field of computer vision and image classification; small perturbations in the data can lead to incorrect classifications by the DNN \cite{DNNAttacks1, DNNAttacks2}. These vulnerabilities raise concerns about the robustness of the machine learning technology that is being adopted and, in some cases, in how safe relying on their predictions could be in high-risk scenarios such as autonomous driving \cite{DNNSelfDriving} and medical diagnosis \cite{DNNMedical}.
By modelling the adversary's behaviour and anticipating these attacks, we can train classifiers that are resilient to such changes in the distribution before they occur.

Game theory provides an effective and popular technique to model adversarial evasion scenarios. These games see one player, the learner, attempt to train a prediction model while another player, the adversary, modifies their data in an attempt to influence particular outcomes from the learner's predictor. See, for example \cite{Adv_Class} for early work in modelling adversarial machine learning. The precise structure of such a model then depends on a number of factors, such as whether the attack occurs either at training time \cite{Bruck_Ext}, or implementation time \cite{Brückner_Scheffer_2011}. Further to this, while some games assume the players act simultaneously \cite{NE_static, Deep_learning_Games}, others allow for sequential play where either the adversary acts first \cite{Liuetall, AdvLeadExt, AdvLeadExtTwo, kantarcıoğlu_xi_clifton_2010} or the learner acts first \cite{taxonomy, Brückner_Scheffer_2011}. We focus our attention on attacks made at implementation time, where the adversary seeks to influence an already established prediction model. Sequential evasion games naturally take on the form of a bilevel optimisation program. In this formulation, the learner, taking the role of the leader in the upper-level, train a prediction function while anticipating how the adversary, taking the role of the follower in the lower-level, will react. The solution to this program then results in a resilient prediction function. Pessimistic bilevel optimisation in particular has proved particularly effective. When multiple optimal solutions are available to the adversary, the pessimistic model, unlike its optimistic variant, assumes no cooperation between the adversary and the learner. This formulation better captures the antagonistic nature of adversarial evasion scenarios \cite{benfield2024classificationstrategicadversarymanipulation, second_paper, Brückner_Scheffer_2011}.


A pessimistic bilevel formulation for evasion attacks was initially proposed in \cite{Brückner_Scheffer_2011} to suit classification tasks, which was later extended in \cite{benfield2024classificationstrategicadversarymanipulation, second_paper} by relaxing assumptions on convexity and uniqueness of solutions, again for classification problems.
In this article, we present a novel pessimistic bilevel program to train resilient predictors against evasion attacks in regression problems. We consider scenarios where we expect adversaries to modify their data in an attempt to influence a particular label from a prediction model.  For example, consider a real estate surveyor who uses a prediction model to evaluate the selling price of houses. A homeowner might provide falsified information about their house, such as it's size or distance from public services, in order to influence a higher valuation from the surveyor. Consider a simple example of two houses, $H_1$ and $H_2$. These houses are identical in every aspect except for their location. While $H_1$ is located just $1$km from the nearest train station, $H_2$ is located further away at $4$km. Consequently, while $H_1$ is evaluated at £$500$k, $H_2$ would receive a considerably lower evaluation due to its inaccessibility to public transport. To maximise the selling price, the owner of $H_2$ might lie about the distance of their house as closer to the train station than it actually is. Decisions like this are simulated by the adversary in the lower-level of our bilevel program. Then, the upper-level trains the surveyor's prediction model while considering these adversarial influences and hence trains a more resilient prediction model. Within our model, we make no assumptions about the convexity of the lower-level problem or the uniqueness of its solution, retaining the benefit provided by the pessimistic formulation.

Further to this, we introduce constraints on the lower-level optimisation problem which restrict the extent to which the adversary can modify their data. Consider again our simple housing market of houses $H_1$ and $H_2$. We established that the owner of $H_2$ can lie about the distance of their house from the nearest train station in order to gain a higher evaluation form the surveyor. Without restrictions on their movement, the owner of $H_2$ would simply state that their house is next to the train station, since a smaller distance leads to a higher evaluation. However, it is incredibility easy to spot this lie due to the large discrepancy between the true distance and the distance stated by the owner. A smaller change in distance, on the other hand, could enable the owner to receive a higher evaluation price while remaining plausibly realistic enough that the surveyor does not notice.
Clearly, an adversary which takes smaller movements better reflects the strategies played by adversaries in the real world. However, without restrictions on their movement, the adversary will construct data which, while optimising their objective function, is not plausible or possibly nonsensical. The constraints we introduce here ensure that the similarity between the modified data and its original value is above some pre-defined threshold. To the best of our knowledge, existing game-theoretic approaches to adversarial regression analysis, see, for example, \cite{adv_reg_1, adv_reg_2, adv_reg_multi}, do not propose a pessimistic bilevel program with lower-level constraints on the adversary's movements that makes no assumptions on convexity or uniqueness of solutions.

Moreover, where existing pessimistic approaches rely on feature maps, such as principal component analysis (PCA), or contextual embeddings to represent the adversary's data, the adversary is allowed to manipulate their data in its original feature space. Consequently, we can view the explicit values of the adversary's data at their solution. These values can give us insight into the strategies played by the adversary and allows us to identify the extent to which each feature needs to be modified in order to adequately influence the prediction model. From this, we can identify which features are most vulnerable. For example, if a certain feature required only a minor perturbation, then we might consider it to be particularly vulnerable to adversarial influence. On the other hand, if a feature required large transformations to trick the learner, then we might consider it to be considerably more resilient to attacks.

The contributions of this article can be summarised as follows. We present a novel pessimistic bilevel optimisation program to train regression prediction functions that are resilient to strategic evasion attempts. This bilevel program, unlike existing approaches to adversarial regression, restricts the adversary's movement through lower-level constraints, while also making no assumptions on the convexity of the lower-level problem, or the uniqueness of its solution. We design experiments to assess the performance of our model and showcase its ability to outperform existing methods. Finally, we demonstrate a key benefit of our model over existing pessimistic approaches. Where existing approaches transform the adversary's data before analysis, removing the ability to analyse the adversary's data at their solution, we keep it in its original feature space. This grants us the ability to infer which features require the greatest transformation in order to effectively influence the learner's prediction function.

The remainder of this article is outlined as follows; we begin in Section \ref{sctn:bilevelModel} by presenting the pessimistic bilevel model with lower-level constraints to train resilient prediction functions in regression analysis before outlining the solution method. Following this, we assess the performance of the bilevel model on two datasets in Section \ref{sctn:regression_experiments} before concluding our findings in Section \ref{sctn:conclusion}.

\section{Training resilient prediction functions}
\label{sctn:bilevelModel}

In this section we present the pessimistic bilevel program with lower-level constraints to model adversarial regression. In particular, we demonstrate how the process of adversarial training can be used to train resilient predictors under the scenario of adversaries attempting to strategically modify their data in an attempt to be assigned incorrect labels from established prediction models. Such scenarios might arise, for example, in property valuation, as outlined in the introduction.
Through suitable choices of objective functions, we construct a constrained pessimistic bilevel program that sees a learner, in the upper-level, train a prediction function while considering how an adversary, in the lower-level, might alter their data towards a target outcome from the prediction model. The adversary is assigned some existing data as a starting point and the distance that the adversary diverges from this point is measured and restricted by lower-level constraints. We demonstrate, through a motivating example, how these constraints ensure that the adversary's data remain plausibly realistic and prevent instances of data losing their intended meaning. In this way, the bilevel model more realistically simulates adversarial transformations, leading to improved performance. To the best of our knowledge, a pessimistic bilevel program with constrained adversarial movement and with no assumptions on convexity or uniqueness of lower-level solution has not previously been proposed for adversarial evasion scenarios in regression analysis.


Let $x \in \mathbb{R}^q$ be a sample (row) of data containing $p \in \mathbb{N}$ features with corresponding label $y \in \mathbb{R}$. For some weights $w \in \mathbb{R}^q$, the learner's prediction function, $\sigma : \mathbb{R} \times \mathbb{R} \rightarrow \mathbb{R}$, as the linear weighted sum of the features:
\begin{equation}
\label{eqn:regressiob prediction}
    \sigma(w,x) := w^Tx.
\end{equation}
This is a common prediction function in linear regression. We then construct a loss function, $\mathcal{L} : \mathbb{R} \times \mathbb{R} \rightarrow \mathbb{R}$, to train the prediction function and identify appropriate weights. As in linear regression, we can achieve this by minimising the squared error between the learner's prediction of $x$ and the label, $y$,
\begin{equation}
\label{eqn:learners loss}
    \mathcal{L}(\sigma(w,x), y) := (\sigma(w,x) - y)^2 = (w^Tx - y)^2.
\end{equation}
The adversary, in the lower-level, transforms some data with the aim of having the learner's prediction function, $\sigma$ output a desired target label. For example, consider again the example of an adversary lying about features of a housing property to influence a higher valuation than deserved. If, for example, the actual valuation price is $y \in \mathbb{R}$, the adversary might set their target, $z \in \mathbb{R}$, as $z = y + \nu$, where $\nu > 0$. The lower-level (adversary's) loss function is then defined as the squared distance between the learner's prediction and the target label:
\begin{equation}
\label{eqn:adversary's regression loss}
    \ell(\sigma(w,x), z) := (\sigma(w,x) - z)^2 = (w^Tx - z)^2.
\end{equation}
With the loss functions established, we now work them into objective functions that accommodate a full dataset. We divide the training data into two sets, a static set as would typically be used in the training of a machine learning predictor, and then a second set of data which can be manipulated by the adversary. Let $D \in \mathbb{R}^{n \times q}$ be the static set of $n \in \mathbb{N}$ instances of data where each $D_i \in \mathbb{R}^q, \; i =1,\dots,n$, is a row vector containing the values of $q \in \mathbb{N}$ features, and let
$\gamma \in \mathbb{R}^{n}$ be the corresponding corresponding collection of labels. Let $X \in \mathbb{R}^{m \times q}$ be the set of $m \in \mathbb{N}$ instances of the same $q$ features which can be manipulated by the adversary with corresponding labels $Y \in \{0,1\}^m$. For convenience, we collapse the adversary's data into a single column vector in order to reduce its dimensionality. Therefore, the adversary's data, $X \in \mathbb{R}^{mq}$ is refined as
\[X :=\begin{pmatrix}
    X_1^T \\
    \vdots \\
    X_m^T
\end{pmatrix},\]
where each $X_i, \; i = 1, \dots, m$, is a row vector of features. For consistency, we also redefine the static data in the same way, $D \in \mathbb{R}^{n q}, D:=(D_1,\dots,D_n)^T$.

The upper level objective of the bilevel program sees the learner minimising their loss over both sets of data. Note that the learner considers the true labels of the adversary's data,
where $F : \mathbb{R}^{p} \times \mathbb{R}^{m p} \rightarrow \mathbb{R}$ is the upper-level objective defined by
\begin{equation}
\label{eqn:learners objective}
    F(w, X) := \frac{1}{n} \sum_{i=1}^n \mathcal{L}(\sigma(w,D_i),\gamma_i) + \frac{1}{m} \sum_{i=1}^m \mathcal{L}(\sigma(w,X_i),Y_i).
\end{equation}
Then, given a set of weights, $w$ of the learner's prediction function, the adversary, in the lower-level seeks to modify their data such that the learner's prediction moves towards their target labels, $Z$. Let
$f : \mathbb{R}^p \times \mathbb{R}^{m p} \rightarrow \mathbb{R}$ be the adversary's objective, which is defined as the sum of the adversary's loss over their data:
\begin{equation}
\label{eqn:adversary regression}
    f(w,X) := \frac{1}{m} \sum_{i = 1}^m \ell(\sigma(w, X_i), Z_i).
\end{equation}



As we have highlighted through a motivating real estate example in Section \ref{sctn:Introduction}, an adversary who takes smaller movements better reflects the strategies played by adversaries in the real world. However, the unconstrained objective functions do not take this under consideration. The optimal adversary under the objective function in \eqref{eqn:adversary regression} will overestimate the abilities of the adversary, leading unrealistic data transformations. In order to construct a realistic adversary within the bilevel program, we introduce a set of lower-level constraints on the adversary's optimization problem which measure the similarity between the true value of the adversary's data and the value seen by the learner. Let $X^0$ be the true (original) value of the adversary's data, and let $g : \mathbb{R}^{mq} \rightarrow \mathbb{R}^m$ be the vector of constraint functions
\begin{equation}
\label{eqn:constraints_array}
    g(X) := \begin{pmatrix}
        g_1(X) \\
        \vdots \\
        g_m(X)
    \end{pmatrix}.
\end{equation}
We define each constraint function component $g_i : \mathbb{R}^q \rightarrow \mathbb{R}$, for each instance of the adversary's data $i=1, \ldots, m$, by 
\begin{equation}
\label{eqn:constraints}
    g_i(X) := \delta - d(X_i, X^0_i),
\end{equation}
where $d : \mathbb{R}^{q} \times \mathbb{R}^{q} \rightarrow \mathbb{R}$ is some similarity function and $\delta$ is the minimum required similarity threshold. The constraint $g(X) \leq 0$ then restrict the adversary to produce data whose similarity is greater than $\delta$.
The work in \cite{second_paper} demonstrated the effectiveness of the cosine similarity to measure the extent to which the adversary has modified their data and so we use the same here,
\begin{equation}
\label{eqn:regression_constraints}
    d(X_i, X^0_i) := \frac{X_i \cdot X_i^0}{\Vert X_i \Vert \Vert X_i^0 \Vert}, \; i = \{1,\dots,m\}.
\end{equation}
Combining the lower-level objective \ref{eqn:adversary regression} with the constraints in \ref{eqn:regression_constraints} gives the adversary's optimisation problem, the solution to which provides data that trick the learner into giving the best possible label assignment while remaining plausibly realistic. Given a set of weights of the learner's prediction function, $w$, we defined $S(w)$ as this set of optimal data, given by
\begin{equation}
\label{eqn:regression solution set}
    S(w) := \argmin_{ X \in \mathbb{R}^{m q}} \left \{ f(w, X) \mid g(X) \leq 0 \right \}.
\end{equation}

Note that the adversary's objective function under the linear regression loss in \eqref{eqn:adversary's regression loss} and prediction function \eqref{eqn:regressiob prediction} possess multiple optimal solutions. This can be directly shown using \cite[Proposition 1]{second_paper} since the prediction function, $\sigma(w,x) = w^Tx$, is a linear combination of $w$ and $x$. Note also that it has already been  demonstrated in \cite[Proposition 2]{second_paper} that the feasible region defined by the constraints in \eqref{eqn:regression_constraints} is non-convex. Therefore, in the case of multiple optimal solutions to the adversary's problem, given by the set $S(w)$, we make the pessimistic assumption that the adversary will not act cooperatively, namely, they will choose the solution which maximises the learner's objective, while the learner seeks to minimise it. The complete bilevel program is then given by
\begin{align}
\label{eqn:bilevel}
    \min_{w \in \mathbb{R}^p} \max_{X \in S(w)} F(w,X).
\end{align}

A solution to the problem described by \eqref{eqn:regression solution set} - \eqref{eqn:bilevel} will comprise the weights of a prediction that has accounted for adversarial manipulation during its training process. Consequently, the prediction function will be more resilient to evasion attacks made after implementation. We say that a set of weights $\bar w$ is a local optimal solution for problem \eqref{eqn:regression solution set} - \eqref{eqn:bilevel} if there is a neighbourhood $W$ of the point such that 
\begin{equation}\label{eq:OptimalConcept}
    \varphi_p(\bar w) \leq \varphi_p(w) \;\, \mbox{ for all }\; w\in W,
\end{equation}
where the $\varphi_p$ denotes the following \textit{two-level value function} (concept first introduced and studied in \cite{dempe2012sensitivity}):
\[
\varphi_{\textup{p}}(w):= \max\{F(w,X)\,|\,X\in S(w)\}.
\]
Note that the the two-level value function $\varphi_{\textup{p}}$ is typically non-convex. Hence, problem \eqref{eqn:regression solution set} - \eqref{eqn:bilevel} is a nonconvex optimization problem. For such problems, algorithms usually only compute stationary points. Therefore, we aim to calculate stationary points for problem \eqref{eqn:regression solution set} - \eqref{eqn:bilevel}. If we have a local optimal solution, $w$, of the bilevel program, then, based on results in \cite{dempe2014necessary,dempe2019two} (and  \cite[Theorem 1]{second_paper} for more precise relevant calculations), under a suitable framework, we can find a Lagrange multiplier vector $(\lambda, \beta, \hat \beta)$ such that the following stationarity conditions are satisfied:
\begin{subequations}
\label{eq:KKT_pbpp_abstract_rerefined}
	\begin{align}
		\label{KTpes6}
        \nabla_w F(w,X) & = 0,\\
        \label{KTpes6.1}
		\nabla_X F(w, X) - \lambda \nabla_X f(w, X) - \nabla g(X)^\top     \beta & = 0,\\
		\label{KTpes7}
		\nabla_X f(w, X) + \nabla g(X)^\top    \hat\beta& = 0,\\
		\label{KTpes8}
		\hat\beta \geq 0, \quad g(X)\leq 0,\quad \hat\beta^\top     g(X)&=0,\\
        \label{KTpes8.1}
      \lambda \geq 0,\;\,  \beta\geq 0,\quad g(X)\leq 0,\quad \beta^\top     g(X)&=0,
 	\end{align}
\end{subequations}
for some data $X$ belonging to the adversary. 
We can solve this system by first transforming it into a system of equations. To simplify the notation, we introduce the  block variables
\[
        z:= \begin{bmatrix}
            w \\
            X
        \end{bmatrix} \in \R^{q +mq} \;\mbox{ and } \;\xi:=\begin{bmatrix}\beta\\\hat\beta\\\lambda\end{bmatrix}\in\R^{2b+1},
\]
as well as the  block functions
\[
	\calG(z)
	:=
	\begin{bmatrix}
		g(X)\\g(X)\\0
	\end{bmatrix}
	\; \mbox{ and }\;
	\calH(z,\xi)
	:=
	\begin{bmatrix}
		\nabla_w F(w,X)\\
		\nabla_X L^\textup{p}_X(z,\xi)\\
		\nabla_X \ell^\textup{p}(z,\xi)
	\end{bmatrix},
\] where $L^\textup{p}_X,\, \ell^\textup{p}\colon(\R^q \times \R^{m q})\times\R^{2b+1}\to\R$ are upper and lower-level Lagrangian type functions: 
\begin{align*}
	& L^\textup{p}_X(z,\xi)
	:=
	F(w,X)-\lambda f(w,X)-\beta^\top g(X), \\
	& \ell^\textup{p}(z,\xi)
	:=
	f(w,X)+\hat\beta^\top g(X).
\end{align*}

Based on this notation, the system described by \eqref{KTpes6}--\eqref{KTpes8.1} can then be restated as 
\begin{equation}\label{eqn:system-Eq}
    \zeta \geq 0, \;\; \calG(z) \leq 0, \;\; \zeta^T \calG(z) = 0, \;\; \calH(z, \xi) = 0,
\end{equation}
which can equivalently be written as the following system of equations
\begin{equation}
\label{eqn:equality system}
\left\{\begin{array}{l}
    \calH(z, \zeta)=0, \\
    \vartheta_{\textup{FB}} \left ( \zeta_i, -\calG_i(z, \zeta) \right )=0, \;i=1, \ldots, 2m. 
\end{array}
\right.
\end{equation}
Here, $\vartheta_{\textup{FB}}$ corresponds to the so-called Fischer-Burmeister function \cite{fischer1992special}, which is defined by
\begin{equation*}
\label{eqn:FB}
    \vartheta_{\text{FB}} (a,b) := \sqrt{a^2 + b^2} - (a+b) \;\mbox{ for }\; (a,b)\in \mathbb{R}^2.
\end{equation*}
Thanks to this function, the second equation in \eqref{eqn:system-Eq} is equivalent to the complementarity conditions in \eqref{KTpes8}--\eqref{KTpes8.1}, written in compact form in \eqref{eqn:system-Eq}, while considering them in pairs. 

Clearly, the stationary conditions \eqref{KTpes6}--\eqref{KTpes8.1} of problem \eqref{eqn:regression solution set} - \eqref{eqn:bilevel} have been written as a system of equations in \eqref{eqn:equality system}. This system is overdetermined, with $m$ more equations than variables, making the Levenberg-Marquardt method, specifically, the global nonsmooth Levenberg–Marquardt method for mixed nonlinear complementarity systems developed in \cite{LevenbergMarquardt}, a suitable choice to solve it. A version of this algorithm which has been appropriately modified for the pessimistic adversarial bilevel program by introducing additional stopping criteria, is given in \cite[Algorithm 2]{second_paper}. We apply this same algorithm here to solve system \eqref{eqn:equality system}, using the derivative formula provided in Appendix \ref{appendix-section},  and extract the solution $z^* = (w^*, X^*)$. This solution comprises the weights, $w^*$, of a prediction function that has accounted for adversarial influence during the training process as well as the corresponding transformed adversary's data, $X^*$. Unlike the framework proposed in the previous work in \cite{second_paper}, it is not required that the adversary's data is embedded ahead of analysis. Consequently, we can use $X^*$ to observe the value of the adversary's solution in its original feature space and investigate the nature of their transformation.

\section{Numerical experiments}
\label{sctn:regression_experiments}

We assess the resilience of the bilevel model to strategic adversarial modifications. To do so, we design experiments on two datasets, the first of which, named \textit{Wine Quality} records 11 features of 4898 bottles of red wine such as acidity and alcohol content. A corresponding quality score between $1-10$ is measured and recorded for each bottle. For this dataset, we consider scenarios where wine producers might lie when reporting the features of the wine or perhaps bribe quality assessors to alter the results in order to appear to be producing wine of a higher quality than they actually are. The second dataset, named \textit{Real Estate}, contains the values of 6 features of 414 houses such as their age and distance to the nearest train station along with the corresponding sale price. We consider scenarios where sellers might lie about features of the house in order to be able to charge a higher price than deserved. Further to this, we use the numerical experiments in this section to explore a particularly useful aspect of the constrained model which, due to the requirement of feature transformations, was not exploitable in previous works. Specifically, we demonstrate how, unlike existing pessimistic bilevel approaches to adversarial training, our models allow us to investigate which features of the adversary's data were changed and hence give us insight into the most vulnerable aspects of the prediction models.

We divide the data into a training, and test set by the ration of $80\%$ and $20\%$ respectively and simulate adversarial attacks to inject the into the test set.
Let $X^{\text{test}} \in \mathbb{R}^{T \times p}$ be the test set containing $T$ samples. Before evaluation, we simulate adversarial influence on the test set by transforming a portion of the instances towards target values $Z^\text{test} = Y^{\text{test}} + \Delta$, defined as a perturbation of the ground truth test labels $Y^\text{test}$, where $\Delta \in \mathbb{R}$ is a perturbation. Let $t <= T$ be the portion of data modified by an adversary and let $I \subset \{1,\dots,T\}$ be the corresponding indexes of the test data which are manipulated. We simulate a range of adversarial abilities. We generate adversarial test data by modifying $X^\text{test}_I$ towards the adversary's target labels. To simulate a range of abilities, we randomly generate each adversary their own similarity threshold from the range $(0.8,1)$. Let $\delta^{\text{test}} \in (-1,1)^t$ be the set of similarity thresholds where $\delta_i \sim U(0.8,1) \; \forall \; i \in I$ and $U$ denotes the uniform distribution. The instances of adversarial test data and then found by solving the problem 
\[X^\text{test}_{i} := \argmin_{x \in \mathbb{R}^{q}} \left \{ \ell \left ( \sigma(w^\text{init}, x \right ), Z^\text{test}_i) \, \left| \, \delta^\text{test}_i - \frac{x \cdot Z^\text{test}_i}{\Vert x \Vert \Vert Z^\text{test}_i \Vert} \leq 0 \right. \right \}, \; \forall i \in I.
\]
where $x^0 \in \mathbb{R}^q$ is the initial values of $x$. We set $T$ to be $10\%$ of the size of the test set. All features and labels are normalised to the range $(0,1)$ and we set the adversarial perturbation to be $\Delta = 2 \text{std}(Y^{\text{test}})$ where $\text{std}(Y^{\text{test}})$ is the standard deviation of the test labels.

\begin{figure}[H]
    \centering
    \includegraphics[scale = 0.38]{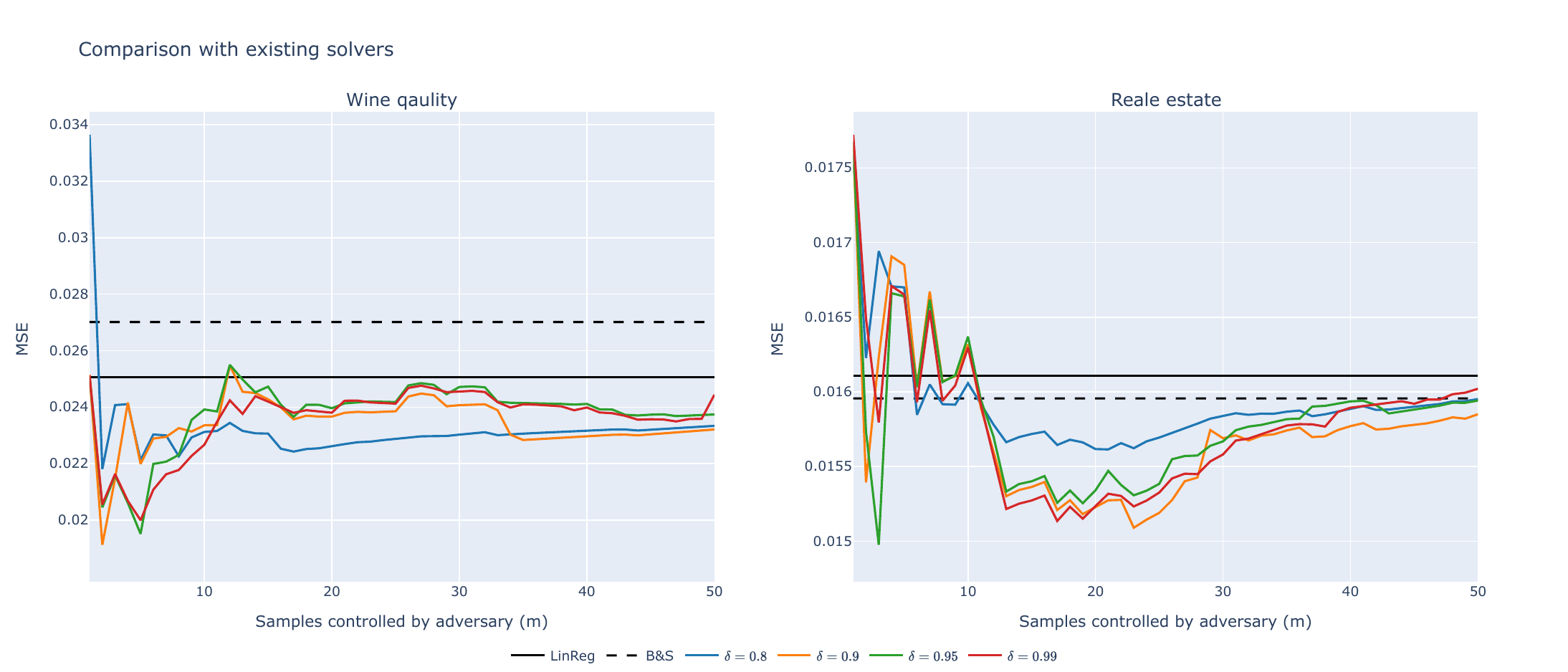}
    \caption{Comparison of the pessimistic bilevel model with existing solvers}
    \label{fig:ms_combined}
\end{figure}

We compare our model to that of a typically trained linear regression predictor, named \textit{LinReg}. While the pessimistic model in \cite{Brückner_Scheffer_2011} was intended for use on classification tasks, its prediction function and loss functions can be easily substituted for linear regression variants which still satisfy their assumptions about strict convexity. We name this \textit{B\&S} and use it as a comparison to the existing pessimistic approach to adversarial regression analysis. The mean square error for these models as well as our model for various values of the adversary's sample size, $m$, and the similarity threshold, $\delta$, are plotted in Figure \ref{fig:ms_combined}. We can see for both datasets a similar trend of an initially high mean square error (MSE) for low values of $m$, before a decrease to an optimal value. Following this, we observe an increase in MSE, before plateauing around an MSE roughly equal to that of the traditional predictor. Both experiments see an optimal MSE when $\delta = 0.95$ and when $m$ is fairly low. For the \textit{wine} dataset, the optimal value falls at $m=1$, while on the \textit{real estate} dataset, we see the optimal value fall at $m=2$. Although, we note that the \textit{real estate} dataset also sees similarly good performance for $m$ in the range of $13 - 23$. In general, we observe a general pattern in that the adversary should be granted enough freedom and influence over the training process to capture the nature of adversarial attacks, while not too much influence that the model becomes over-estimates the power of the adversary and hence suffers in performance.

\begin{figure}[H]
    \centering
    \includegraphics[scale = 0.38]{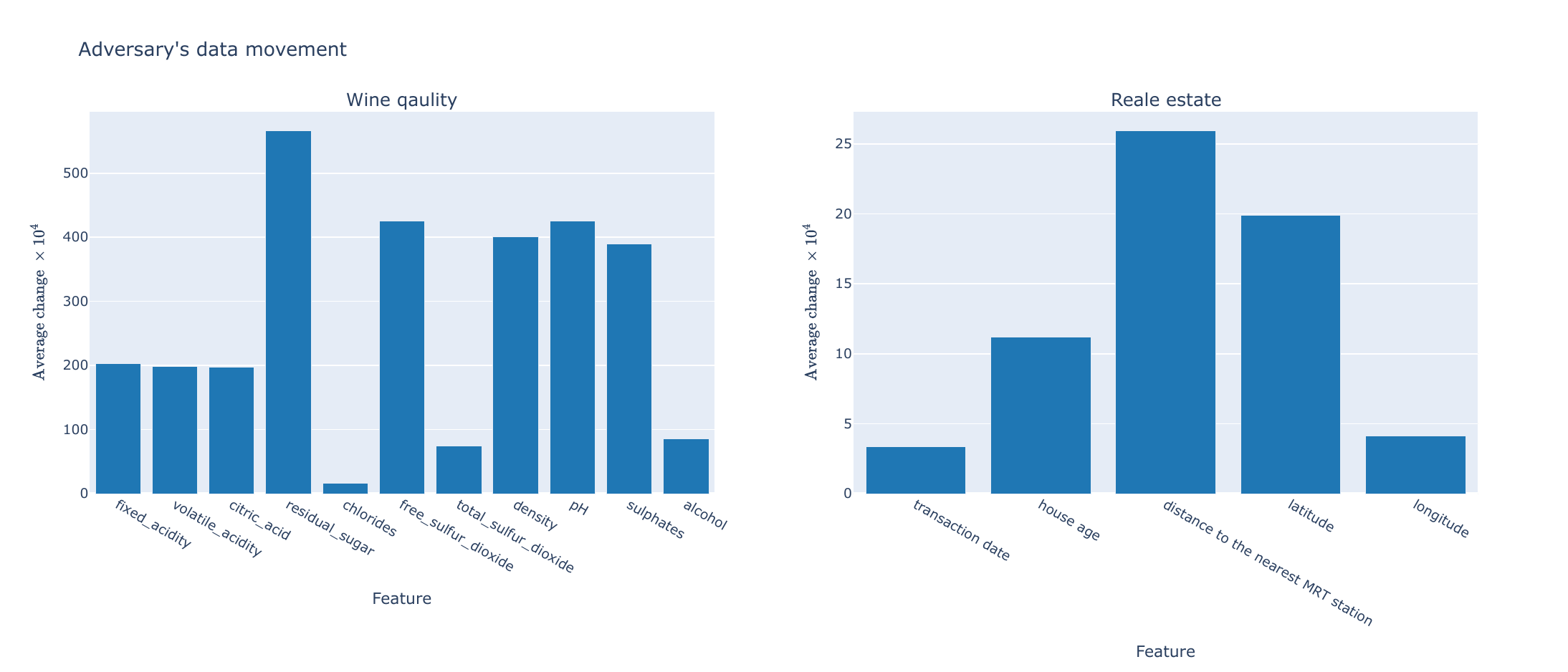}
    \caption{Feature modifications made by the adversary to trick the learner}
    \label{fig:feature_combined}
\end{figure}

We now investigate the movement of each feature by the adversary during the training process to gain insight into which features are most vulnerable. For each feature (column), we calculate the average absolute distance between the adversary's solution $X^{*}$ and the original position $X^0$,
\[\frac{1}{m} \sum_{j=1}^m \left \vert X^*_{ij} - X^0_{ij} \right \vert , \quad i = 1,\dots,m.\]
Under this measure, a larger distance implies that the adversary needed to transform the data by a larger extent in order to sufficiently fool the leaner's prediction model. We plot the average movements for each feature and for each dataset in Figure \ref{fig:feature_combined}. For the \textit{Wine quality}, we see the most movement in residual sugar followed by PH and sulfur dioxide. These results give the insight into how much an adversary must modify each feature in order to fool the learner. It is clear, for example, that the alcohol content must be modified considerably to generate the most effective wine. While chlorides, on the other hand, required little change to achieve a value sufficient to fool the learner's predictor. Chlorides, therefore, might be considered a particularly vulnerable feature that could easily be exploited and perhaps should be treated with more scrutiny. In the \textit{real estate} data, we see the distance to the nearest MRT station as the most resilient feature, requiring relatively substantial change to affect the learner's predictor. Meanwhile, the transaction data and longitude appear to be the easiest to exploit.

It is clear from the experimental results that the pessimistic bilevel model with constrained adversarial movement provides an effective method of accounting for adversarial manipulation during the training process of regression scenarios. The lower-level successfully anticipates the movements of an adversary, which leads to improved performance by the prediction model trained in the upper-level. We see a significant change in performance as we increase the number of instances available to be manipulated by the adversary. In particular, we note that a balance needs to be struck between allowing the adversary enough instances that they have sufficient influence over then training process, while not too many that the model overestimates the influence of the adversary. Finally, we demonstrated how, unlike existing works, we could investigate the values of the adversary's data at the solution. This allowed us to identify which features were most vulnerable to exploitation.

\section{Conclusion}
\label{sctn:conclusion}

In this article, we proposed a pessimistic bilevel program to model adversarial evasion in regression scenarios. In particular, by anticipating adversarial movement in the lower-level, we can train prediction functions that are resilient to adversaries who attempt to influence particular outcomes from a prediction model by strategically transforming their data. For example, in the context of property valuation, an adversary might purposefully provide falsified information to achieve higher valuations and hence charge a higher selling price. While there exist pessimistic bilevel approaches to adversarial evasion scenarios, these models exploit strong assumptions about the convexity of the adversary's problem and the uniqueness of their solution to reformulate the program into its optimistic variant, which, while easier to solve, oversimplifies the capabilities of the adversary. Since our model makes no such assumptions, we retain the pessimistic aspect of the bilevel model, allowing us to accurately capture the antagonistic nature of these scenarios. Furthermore, with the introduction of lower-level constraints, which restrict the adversary's movement, preventing nonsensical transformations and ensuring the adversary's data remains plausibly realistic, leading to more accurate prediction functions.

We simulated adversarial influence to create test sets for wine quality appraisal and real estate pricing. Numerical experiments demonstrated the ability of the bilevel model to train resilient predictors that provided improved performance over existing methods. Further to this, we investigated varying the number of samples available to the adversary and observed a clear pattern in the form of a trade-off between allowing the adversary enough freedom and influence to generate sufficient movement that impacts the training process, while not too much such that the adversary provides too much influence over the training process, leading to overly pessimistic predictors. Additionally, we investigated a feature of the pessimistic bilevel model which previous works were not able to exploit. Specifically, since we do not transform the adversary's data before analysis, we were able to observe the adversary's solution in its original feature space and measure the extent to which the adversary modified each feature. From this we could identify which features were most vulnerable to attacks.


\appendix

\section{Derivatives for the leader and follower}\label{appendix-section}
Let $x \in \mathbb{R}^q$ be a sample of data with corresponding label $y \in \mathbb{R}$ and let $w \in \mathbb{R}^q$ be the weights of the learner's prediction function, $\sigma : \mathbb{R} \times \mathbb{R} \rightarrow \mathbb{R}$ defined as in \eqref{eqn:regressiob prediction}.
We define the upper-level (learner's) loss function $\mathcal{L} : \mathbb{R} \times \mathbb{R} \rightarrow \mathbb{R}$ as it is in \eqref{eqn:learners loss}.
Let $z \in \mathbb{R}$ be the adversary's target label.

Let $D \in \mathbb{R}^{m q}$ be the static data with corresponding labels $\gamma \in \mathbb{R}^n$ and let $X \in \mathbb{R}^{m q}$ be the adversary's data with corresponding labels $Y \in \mathbb{R}^m$. The upper-level objective $F : \mathbb{R}^q \times \mathbb{R}^{m q} \rightarrow \mathbb{R}$ is defined as it is in \eqref{eqn:learners objective}.
The derivative of the upper-level (leader's) loss with respect to the learner's weights is given by,
\[\nabla_w F (w,X) = \frac{2}{n} D^T(w^TD - \gamma) + \frac{2}{m} X^T(w^TX - Y) + \frac{2}{\rho} w .\]
The second derivative of the upper-level loss with respect to the learner's weights is given by
\[\nabla^2_{ww} F (w,X) = \frac{2}{n} D^TD + \frac{2}{m} X^TX + \frac{2}{\rho}\]
The derivative of the upper-level objective function with respect to the adversary's data is given by
\begin{equation*}
\label{eqn:upper_level_derivative}
    \frac{\partial F}{\partial X_{ij}} (w,X) := \frac{1}{m} \sum_{k=1}^m \frac{\partial \mathcal{L}}{\partial X_{ij}} (\sigma(w,X_k),Y_k) = \frac{1}{m} \frac{\partial \mathcal{L}}{\partial X_{ij}} (\sigma(w,X_i),Y_i),
\end{equation*}
where
\[\frac{\partial \mathcal{L}(w,X)}{\partial X_{ij}} = \frac{2}{m} w_j (w^TX_i - y_i), \quad i = 1,\dots,m, \; j = 1,\dots,q.\]
The second derivative with respect to the adversary's data is given by the following cases,
\[\frac{\partial^2 \mathcal{L}}{\partial X_{ij} \partial X_{kl}} = \begin{cases}
    \frac{2}{m} w_j w_l & i = k \\
    0 & \text{otherwise}.
\end{cases}\]

The derivative of the upper-level objective function with respect to the adversary's data is given by
\begin{equation*}
\label{eqn:lower_level_derivative}
    \frac{\partial f}{\partial X_{ij}} (w,X) = \frac{1}{m} \sum_{i=k}^m \frac{\partial \ell}{\partial X_{ij}} (\sigma(w, X_k), Y_k) = \frac{1}{m} \frac{\partial \ell}{\partial X_{ij}} (\sigma(w, X_k), Y_k).
\end{equation*}
where the lower-level (adversary's) loss function $\ell : \mathbb{R} \times \mathbb{R} \rightarrow \mathbb{R}$ is defined as it is in \eqref{eqn:adversary's regression loss}. Note that the adversary's loss function can be expressed in terms of the leader's loss function,
\[\ell(\sigma(w,x), z) = \mathcal{L}(\sigma(w,x),z),\]
for some weights $w \in \mathbb{R}^q$, data $x \in \mathbb{R}^q$ and target label $z \in \mathbb{R}$. Therefore, we can express the derivative of the lower-level objective function with respect to the adversary's data as follows
\[\frac{\partial f}{\partial X_{ij}} (w, X) = \frac{1}{m} \frac{\partial \mathcal{L}}{\partial X_{ij}} (\sigma(w, X_i), Z_i) \quad i = 1,\dots,m, \; j = 1,\dots,q.\]
The second derivative with respect to the adversary's data is then given as
\[
\frac{\partial^2 f}{\partial X_{ij} \partial X_{kc}} (w, X) = \frac{1}{m} \frac{\partial^2 \mathcal{L}}{\partial X_{ij} \partial X_{jk}} (\sigma(w, X_i), Z_i) \quad i,k = 1,\dots,m, \; j,c = 1,\dots,q.
\]
Finally, the derivative with respect to the learner's weights and the adversary's data is given by
\[
\frac{\partial^2 f}{\partial w_i \partial X_{jk}} (w, X) = \frac{1}{m} \frac{\partial^2 \mathcal{L}}{\partial w_i \partial X_{jk}} (\sigma(w, X_i), Z_i) \quad i,j = 1,\dots,m, \; k = 1,\dots,q.
\]

Let $X^0 \in \mathbb{R}^{m q}$ be the start point of the adversary's data and let the lower-level constraints $g : \mathbb{R}^{m q} \rightarrow (-1,1)$ be defined as in \eqref{eqn:constraints_array}, where each constraint function $g_i(X) : \mathbb{R}^q \rightarrow \mathbb{R}, \; i = 1,\dots,m$ measures the cosine similarity between the adversary's data and its original position, as given by \eqref{eqn:regression_constraints},
where $\delta \in \mathbb{R}$ is the similarity threshold. The derivative of the constraints with respect to the classifier weights is $0$. The derivative with respect to the adversary's data is obtained as
\begin{equation*}
\label{eqn:constraints_derivative}
    \frac{\partial g_i(X)}{\partial X_{jk}} = \begin{cases}
    \frac{X_{ik}^0}{\Vert X_i \Vert \cdot \Vert X_i^0 \Vert} - d(X_i, X_i^0) \frac{X_{ik}}{\Vert X_i \Vert ^2} & i = j, \\
    0 & i \neq j.
\end{cases}
\end{equation*}
The second derivative with respect to the adversary's data is given by the cases
\begin{equation*}
\label{eqn:constraints_second_derivative}
    \frac{\partial^2 g_i(X)}{\partial X_{jk} \partial X_{lc}} = \begin{cases}
    \frac{X_{ic} X_{ik}^0 + X_{ik} X_{ic}^0}{\Vert X_i \Vert ^ 3 \Vert X_i^0 \Vert} - \frac{3 X_{ik} X_{ic} d(X_i, X_i^0)}{\Vert X_i \Vert ^4} & i = j = l,\;\, k \neq c, \\\\
    \frac{2 X_{ik} X_{ik}^0}{\Vert X_i \Vert ^ 3 \Vert X_i^0 \Vert} - \frac{3 X_{ik}^2 d(X_i, X_i^0)}{\Vert X_i \Vert ^4} + \frac{d(X_i, X_i^0)}{\Vert X_i \Vert ^2} & i = j = l,\;\, k = c, \\\\
    0 & \text{otherwise}.
\end{cases}
\end{equation*}


\section*{References}

\begin{thebibliography}{10}
\expandafter\ifx\csname url\endcsname\relax
  \def\url#1{\texttt{#1}}\fi
\expandafter\ifx\csname doi\endcsname\relax
  \def\doi#1{\burlalt{doi:#1}{http://dx.doi.org/#1}}\fi
\expandafter\ifx\csname urlprefix\endcsname\relax\def\urlprefix{URL }\fi
\expandafter\ifx\csname href\endcsname\relax
  \def\href#1#2{#2}\fi
\expandafter\ifx\csname burlalt\endcsname\relax
  \def\burlalt#1#2{\href{#2}{#1}}\fi



\bibitem{benfield2024classificationstrategicadversarymanipulation}
David Benfield, Stefano Coniglio, Martin Kunc, Phan Tu Vuong and Alain Zemkoho.
\newblock Classification under strategic adversary manipulation using pessimistic bilevel optimisation. (2024)
\newblock arXiv: https://arxiv.org/abs/2410.20284.

\bibitem{Malware_Detection}
Battista Biggio, Igino Corona, Davide Maiorca, Blaine Nelson, Nedim ˇSrndi´c, Pavel Laskov, Giorgio
Giacinto, and Fabio Roli.
\newblock Evasion Attacks against Machine Learning at Test Time.
\newblock {\em Machine
Learning and Knowledge Discovery in Databases.} Springer Berlin Heidelberg, (2013), p. 387-402.

\bibitem{NE_static}
Michael Br¨uckner and Tobias Scheffer.
\newblock Nash Equilibria of Static Prediction Games.
\newblock {\em Proceed-
ings of the 22nd International Conference on Neural Information Processing Systems.} NIPS’09, (2009).

\bibitem{Brückner_Scheffer_2011}
Michael Br¨uckner and Tobias Scheffer.
\newblock Stackelberg games for adversarial prediction problems.
\newblock {\em Proceedings of the 17th ACM SIGKDD international conference on Knowledge discovery and data
mining - KDD ’11.} ACM Press, (2011).


\bibitem{AdvLeadExtTwo}
Aneesh Sreevallabh Chivukula and Wei Liu.
\newblock Adversarial learning games with deep learning models.
\newblock {\em 2017 International Joint Conference on Neural Networks.} IJCNN, (2017).


\bibitem{adv_deep_learning_survey}
Joana C. Costa, Tiago Roxo, Hugo Proen¸ca, and Pedro Ricardo Morais In´acio.
\newblock How Deep Learning Sees the World: A Survey on Adversarial Attacks \&
Defense.
\newblock {\em IEEE Access.} (2024).


\bibitem{Adv_Class}
Nilesh Dalvi, Pedro Domingos, Mausam, Sumit Sanghai, and Deepak Verma.
\newblock Adversarial Classification.
\newblock {\em Proceedings of the Tenth ACM SIGKDD International Conference on Knowledge Discovery and Data Mining.} KDD ’04. (2004).

\bibitem{GameTheorySurvey}
Prithviraj Dasgupta and Joseph B. Collins.
\newblock A Survey of Game Theoretic Approaches for Adversarial Machine Learning in Cybersecurity Task. (2019)
\newblock {\em AI Mag.} 40.2 (2019) pp. 31–43.


\bibitem{dempe2014necessary}
Stephan Dempe, Boris S Mordukhovich, and Alain B Zemkoho.
\newblock Necessary optimality conditions in pessimistic bilevel programming.
\newblock {\em Optimization.} 63.4. (2014). p. 505-533.


\bibitem{dempe2012sensitivity}
Stephan Dempe, Boris S Mordukhovich, and Alain B Zemkoho.
\newblock Sensitivity analysis for two-level
value functions with applications to bilevel programming.
\newblock {\em SIAM Journal on Optimization.} 22.4. (2012). p. 1309-1343.


\bibitem{DNNSelfDriving}
Kevin Eykholt, Ivan Evtimov, Earlence Fernandes, Bo Li, Amir Rahmati, Chaowei Xiao, Atul
Prakash, Tadayoshi Kohno, and Dawn Song.
\newblock Robust Physical-World Attacks on Deep Learning Visual Classification.
\newblock {\em Proc. 2018 IEEE/CVF Conference on Computer Vision and Pattern Recognition.} (2018). p. 1625-1634.


\bibitem{DNNMedical}
Samuel Finlayson et a.
\newblock Adversarial Attacks Against Medical Deep Learning Systems. (2019)
\newblock arXiv: https://arxiv.org/abs/1804.05296.


\bibitem{fischer1992special}
Andreas Fische.
\newblock A special Newton-type optimization method.
\newblock {\em Optimization.} (1992). p. 269-284.




\bibitem{DNNAttacks2}
Ian J. Goodfellow, Jonathon Shlens, and Christian Szegedy.
\newblock Explaining and Harnessing Adversarial
Examples. (2015)
\newblock {\em Proc. 2015 International Conference on Learning Representations.} (2015).

\bibitem{INFORMS_defense}
Chan Y. Han, Brian J. Lunday, and Matthew J. Robbins.
\newblock A Game Theoretic Model for the
Optimal Location of Integrated Air Defense System Missile Batteries.
\newblock {\em INFORMS Journal on
Computing.} 28.3 (2016), pp. 405–416.

\bibitem{taxonomy}
Ling Huang, Anthony D. Joseph, Blaine Nelson, Benjamin I.P. Rubinstein, and J. D. Tygar.
\newblock Adversarial Machine Learning.
\newblock {\em Proceedings of the 4th ACM Workshop on
Security and Artificial Intelligence.} Association for Computing
Machinery (2011) pp. 43-58.


\bibitem{LevenbergMarquardt}
Lateef O Jolaoso, Patrick Mehlitz, and Alain B Zemkoho.
\newblock A fresh look at nonsmooth Levenberg–Marquardt methods with applications to bilevel optimization.
\newblock {\em Optimization.} 74.12 (2025) pp. 2745-2792.


\bibitem{kantarcıoğlu_xi_clifton_2010}
Murat Kantarcıo˘glu, Bowei Xi, and Chris Clifton.
\newblock Classifier evaluation and attribute selection against active adversaries.
\newblock {\em Data Mining and Knowledge Discovery.} 22.1-2 (2010) pp. 291-335.


\bibitem{Liuetall}
Wei Liu and Sanjay Chawla.
\newblock A Game Theoretical Model for Adversarial Learning.
\newblock {\em Proc. 2009
IEEE International Conference on Data Mining Workshops.} (2009) pp. 25-30.


\bibitem{AdvLeadExt}
Wei Liu, Sanjay Chawla, James Bailey, Christopher Leckie, and Kotagiri Ramamohanarao.
\newblock An Efficient Adversarial Learning Strategy for Constructing Robust Classification Boundaries.
\newblock {\em AI 2012: Advances in Artificial Intelligence.} (2012) pp. 469-660.


\bibitem{Bruck_Ext}
Shike Mei and Xiaojin Zhu.
\newblock Using machine teaching to identify optimal training-set attacks on
machine learner.
\newblock {\em Proc. Twenty-Ninth AAAI Conference on Artificial Intelligence.} AAAI Press (2015) pp. 2871-2877.





\bibitem{Deep_learning_Games}
Dale Schuurmans and Martin Zinkevich.
\newblock Deep Learning Games.
\newblock {\em Proceedings of the 30th International Conference on Neural Information Processing Systems.} NIPS'16 (2016) pp. 1686-1694.


\bibitem{Network_Intrusion}
Robin Sommer and Vern Paxson.
\newblock Outside the Closed World: On Using Machine Learning for Network Intrusion Detection.
\newblock {\em Proc. 2010 IEEE Symposium on Security and Privacy.} (2010) pp. 305-316.



\bibitem{adv_book_2023}
Aneesh Sreevallabh Chivukula, Xinghao Yang, Bo Liu, Wei Liu, and Wanlei Zhou.
\newblock Adversarial Machine Learning: Attack Surfaces, Defence Mech-
anisms, Learning Theories in Artificial Intelligence.
\newblock {\em Proc. 2010 IEEE Symposium on Security and Privacy.} Springer (2023).



\bibitem{DNNAttacks1}
Christian Szegedy, Wojciech Zaremba, Ilya Sutskever, Joan Bruna, Dumitru Erhan, Ian Goodfellow,
and Rob Fergus.
\newblock Intriguing properties of neural networks. (2014)
\newblock {\em Proc. 2014 International Conference
on Learning Representations.} (2014).





\bibitem{INFORMS_credit_card_fraud_2}
Shixiang Zhu, Henry Shaowu Yuchi, Minghe Zhang, and Yao Xie.
\newblock Sequential Adversarial Anomaly Detection for One-Class Event Data.
\newblock {\em INFORMS Journal on Data Science.} 2.1 (2023). pp. 45-59.


\bibitem{second_paper}
David Benfield, Stefano Coniglio, Vuong Phan and Alain Zemkoho.
\newblock Facing Adversarial Data Manipulation via
Constrained Pessimistic Bilevel
Optimization. (2025)
\newblock arXiv: https://arxiv.org/abs/2510.03254.



\bibitem{adv_reg_1}
Scott Alfeld, Xiaojin Zhu, and Paul Barford.
\newblock Data poisoning attacks against autoregressive model.
\newblock {\em Proc. of the Thirtieth AAAI Conference on Artificial Intelligence.}  AAAI Press (2016) pp. 1452–1458

\bibitem{adv_reg_2}
Michael Großhans, Christoph Sawade, Michael Br¨uckner, and Tobias Scheffer.
\newblock Bayesian Games
for Adversarial Regression Problems.
\newblock {\em Proc. of the 30th International Conference on
Machine Learning.}  Proceedings of Machine
Learning Research (2013) pp. 55–63


\bibitem{adv_reg_multi}
Liang Tong, Sixie Yu, Scott Alfeld, and yevgeniy vorobeychik.
\newblock Adversarial Regression with Multiple Learners.
\newblock {\em Proc. of the 35th International Conference on Machine Learning.}  Proceedings of Machine
Learning Research (2018) pp. 4946–4954


\bibitem{dempe2019two}
Stephan Dempe, Boris S Mordukhovich, and Alain B Zemkoho.
\newblock Two-level value function approach to non-smooth optimistic and pessimistic bilevel programs.
\newblock {\em Optimization.}  68.2-3 (2019)
pp. 433–455.

\end{thebibliography}
\end{document}